\documentclass[10pt, conference]{IEEEtran}

\usepackage{amsmath,amssymb,amsfonts}
\usepackage{graphicx}
\usepackage{pifont}

\newcommand{\DisableFootNotes}{%
  \renewcommand{\footnote}[2][]{\relax}
}

\DisableFootNotes

\usepackage{amssymb}
\usepackage{amsthm}
\usepackage{xcolor}
\usepackage[figuresright]{rotating}

\usepackage{graphicx}
\graphicspath{{/}{fig/}}

\usepackage{comment}
\usepackage{array}
\usepackage{textcomp}
\usepackage{gensymb}
\usepackage{xcolor}
\usepackage{multirow}
\usepackage{booktabs}
\usepackage{enumitem}

\usepackage{mathtools}
\usepackage{breqn}
\usepackage{float}
\usepackage[noend]{algpseudocode}
\usepackage[vlined, ruled, shortend]{algorithm2e}

\usepackage{caption}
\usepackage{subcaption}

\usepackage{mathtools}
\usepackage{float}
\usepackage{hyperref}

\usepackage{pgfplots}
\pgfplotsset{compat=1.7}
\usepgfplotslibrary{groupplots}

\usepackage{multirow,tabularx}
\usepackage{flushend}

\usepackage{makecell}

\newcommand{\RN}[1]{%
  \textup{\uppercase\expandafter{\romannumeral#1}}%
}

\newcommand{\RomanNumeralCaps}[1]
    {\MakeUppercase{\romannumeral #1}}
\usepackage{xcolor}
\colorlet{orange_maxim}{green!10!orange!90!}

\usepackage[capitalise]{cleveref}

\newlength\figureheight
\newlength\figurewidth
\usetikzlibrary{shapes.misc}

\tikzset{cross/.style={cross out, draw=black, minimum size=2*(#1-\pgflinewidth), inner sep=0pt, outer sep=0pt},
cross/.default={1pt}}

\renewcommand{\baselinestretch}{0.985}

\IEEEaftertitletext{\vspace{-1.5\baselineskip}}

\title{
   A Benchmark for Multi-Modal Lidar SLAM with Ground Truth in GNSS-Denied Environments
}

\author{
    \IEEEauthorblockN{
        \vspace{1em}
        Ha Sier\IEEEauthorrefmark{2}\IEEEauthorrefmark{1}\textsuperscript{\textsection},
        Li Qingqing\IEEEauthorrefmark{2}\textsuperscript{\textsection},
        Yu Xianjia\IEEEauthorrefmark{2}, 
        Jorge Pe\~na Queralta\IEEEauthorrefmark{2},
        Zhuo Zou\IEEEauthorrefmark{1},
        Tomi Westerlund\IEEEauthorrefmark{2} \\[+0.5em]
    }
    \IEEEauthorblockA{
        \normalsize
        \IEEEauthorrefmark{2}\href{https://tiers.utu.fi}{Turku Intelligent Embedded and Robotic Systems (TIERS) Lab, University of Turku, Finland}.\\
         \IEEEauthorrefmark{1} School of Information Science and Technology, Fudan Universtiy, China \\
        Emails: \IEEEauthorrefmark{2}\{sierha, qingqli, xianjia.yu, jopequ, tovewe\}@utu.fi 
        \IEEEauthorrefmark{1}\{zhuo\}@fudan.edu.cn,\\[+6pt] 
    }
}
  
\begin{document}

\maketitle
\begingroup\renewcommand\thefootnote{\textsection}
\footnotetext{These authors have contributed equally to this manuscript.}
\endgroup
\thispagestyle{empty}
\pagestyle{empty}


\begin{abstract}
    Lidar-based simultaneous localization and mapping (SLAM) approaches have obtained considerable success in autonomous robotic systems. This is in part owing to the high-accuracy of robust SLAM algorithms and the emergence of new and lower-cost lidar products. This study benchmarks current state-of-the-art lidar SLAM algorithms with a multi-modal lidar sensor setup showcasing diverse scanning modalities (spinning and solid-state) and sensing technologies, and lidar cameras, mounted on a mobile sensing and computing platform. We extend our previous multi-modal multi-lidar dataset with additional sequences and new sources of ground truth data. Specifically, we propose a new multi-modal multi-lidar SLAM-assisted and ICP-based sensor fusion method for generating ground truth maps. With these maps, we then match real-time pointcloud data using a natural distribution transform (NDT) method to obtain the ground truth with full 6 DOF pose estimation. This novel ground truth data leverages high-resolution spinning and solid-state lidars. We also include new open road sequences with GNSS-RTK data and additional indoor sequences with motion capture (MOCAP) ground truth, complementing the previous forest sequences with MOCAP data. We perform an analysis of the positioning accuracy achieved with ten different SLAM algorithm and lidar combinations. We also report the resource utilization in four different computational platforms and a total of five settings (Intel and Jetson ARM CPUs). Our experimental results show that current state-of-the-art lidar SLAM algorithms perform very differently for different types of sensors. More results, code, and the dataset can be found at:  \href{https://github.com/TIERS/tiers-lidars-dataset-enhanced}{github.com/TIERS/tiers-lidars-dataset-enhanced.}
\end{abstract}


\IEEEpeerreviewmaketitle

  
\section{Introduction}\label{sec:introduction}

Lidar sensors have been adopted as the core perception sensor in many applications, from self-driving cars~\cite{li2020multi} to unmanned aerial vehicles~\cite{varney2020dales}, including forest surveying and industrial digital twins~\cite{yang2020individual}. 
High resolution spinning lidars enable a high-degree of awareness from the surrounding environments. More dense 3D pointclouds and maps are in creasing demand to support the next wave of ubiquitous autonomous systems as well as more detailed digital twins across industries. However, higher angular resolution comes at increased cost in analog lidars requiring a higher number of laser beams or a more compact electronics and optics solution. New solid-state and other digital lidars are paving the way to cheaper and more widespread 3D lidar sensors capable of dense environment mapping~\cite{van2021solid, qingqing2021adaptive, li2021towards, queralta2020vio}.

\begin{figure}
    \centering 
    \begin{subfigure}{.49\textwidth}
        \centering
        \includegraphics[width=\textwidth]{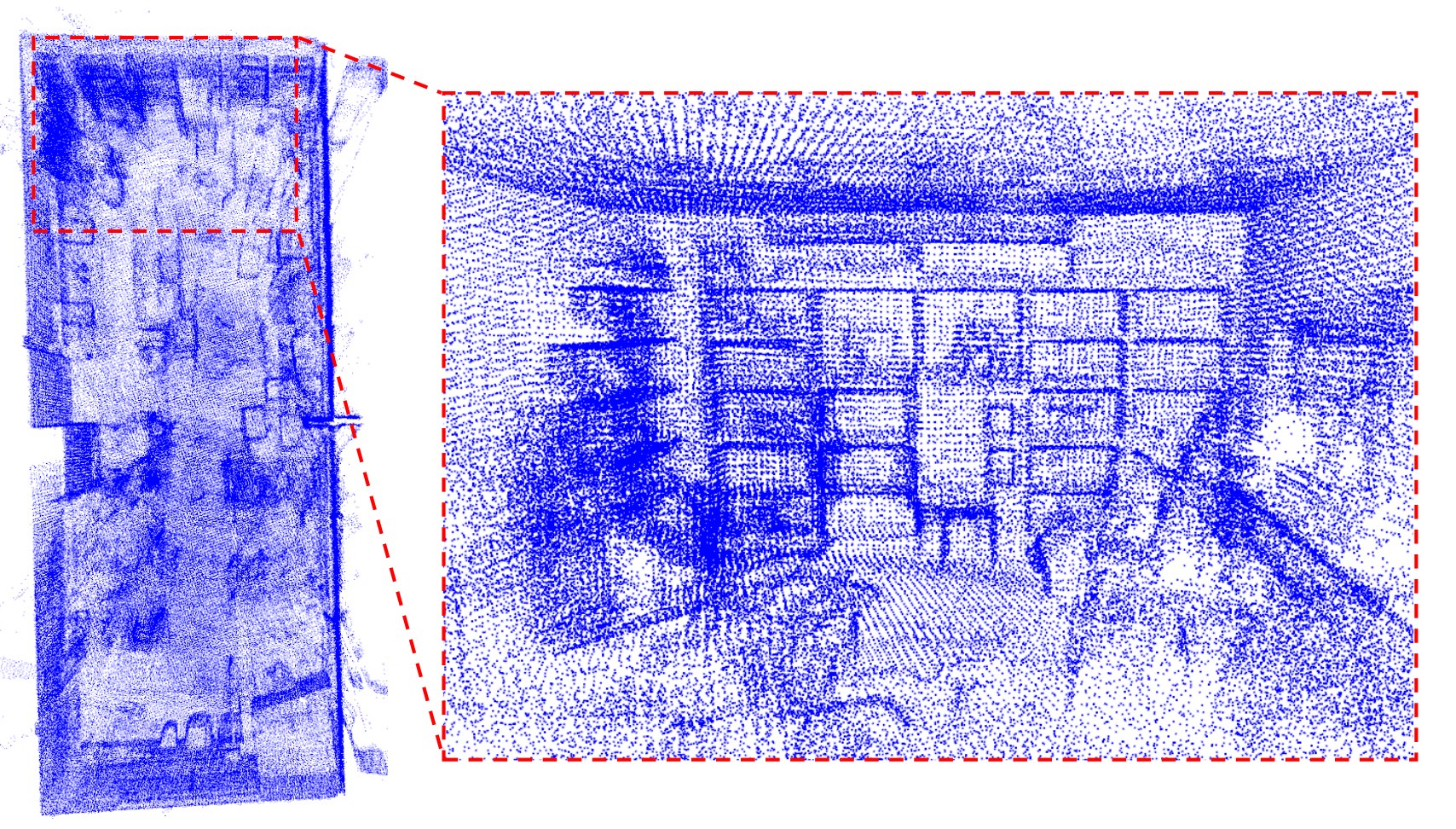} 
        \caption{Ground truth map for one of the indoor sequences generated based on the proposed approach (SLAM-assisted ICP-based prior map). This enables benchmarking of lidar odometry and mapping algorithms in larger environments where a motion capture system or similar is not available, with significantly higher accuracy than GNSS/RTK solutions.} 
        \label{fig:ground_ground_map}
    \end{subfigure}
    
    \vspace{.42em}
    \begin{subfigure}{0.46\textwidth}
        \centering   
          \includegraphics[trim={0cm 0cm 3cm 0cm}, width=\textwidth]{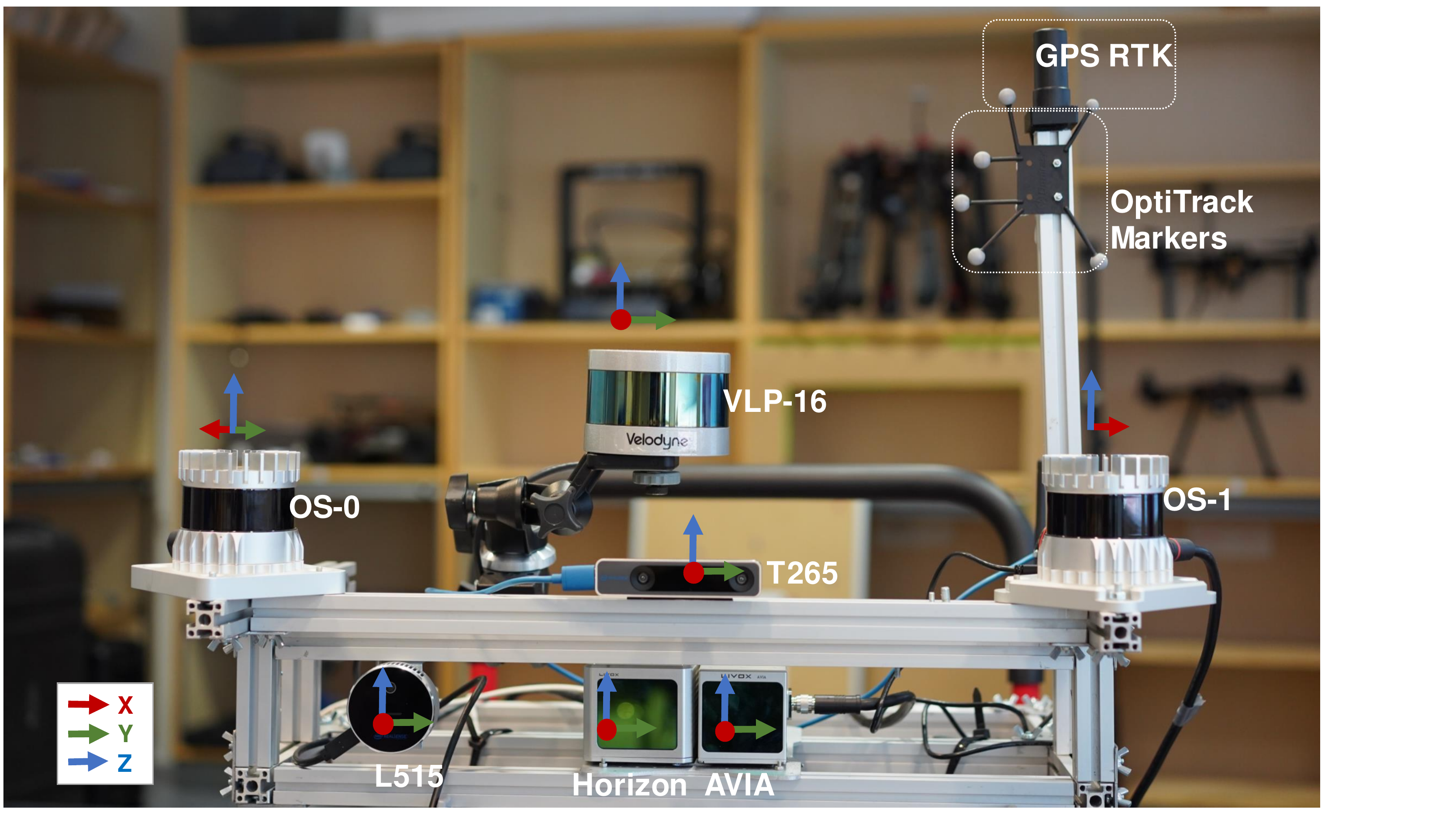}
        \caption{Front view of the multi-modal data acquisition system. Next to each sensor, we show the individual coordinate frames.
        } 
        \label{fig:hardware_cfg}
    \end{subfigure}
    \caption{Multi-modal lidar data acquisition platform and samples from maps obtained in the different environments included in the dataset.}
\end{figure} 


So-called solid-state lidars overcome some of the challenges of spinning lidars in terms of cost and resolution, but introduce some new limitations in terms of a relatively small field of view (FoV)~\cite{lin2020loamlivox, li2021towards}. Indeed, these lidars provide more sensing range at significantly lower cost~\cite{li2022multi}. Other limitations that affect traditional approaches to lidar data processing include irregular scanning patterns or increased motion blur.

Despite their increasing popularity, few works have benchmarked the performance of both spinning lidar and solid-state lidar in diverse environments, which limits the development of more general-purpose lidar-based SLAM algorithms~\cite{li2022multi}. To bridge the gap in the literature, we present a benchmark that compares different modality lidars (spinning, solid-state) in diverse environments, including indoor offices, long corridors, halls, forests, and open roads. To allow for more accurate and fair comparison, we introduce a new method for ground truth generation in larger indoor spaces (see Fig.~\ref{fig:ground_ground_map}). This enhanced ground truth enables significantly higher degree of quantitative benchmarking and comparison with respect to our previous work~\cite{li2022multi}. We hope for the extended dataset and ground truth labels, as well as more detailed data, to provide a performance reference for multi-modal lidar sensors in both structured and unstructured environments to both academia and industry.


In summary, this work evaluates state-of-the-art SLAM algorithms with a multi-modal multi-lidar platform as an extension of our previous work~\cite{li2022multi}. The main contributions of this work are as follows:
\begin{enumerate}
    \item a ground truth trajectory generation method for environments where MOCAP or GNSS/RTK are unavailable that leverages the multi-modality of the data acquisition platform and high-resolution sensors;
    \item a new dataset with data from 5 different lidar sensors, one lidar camera, and one stereo fisheye cameras in a variety of environments as illustrated in Fig.~\ref{fig:hardware_cfg}. Ground truth data is provided for all sequences; 
    \item the benchmarking of ten state-of-the-art filter-based and optimization-based SLAM methods on our proposed dataset in terms of the accuracy of odometry, memory and computing resource consumption. The results indicate the limitations of current SLAM algorithms and potential future research directions.
    
\end{enumerate}%

The structure of the paper is as follows. Section II surveys recent progress in SLAM and existing lidar-based SLAM benchmarks. Section III provides an overview of the configuration of the proposed sensor system. Section IV offers the detailed benchmark and ground truth generation methodology. 
Section V concludes the study and suggests future work. 

 \begin{table*}[ht]  
 \centering
    \caption{Sensor specification for the presented dataset. Angular resolution is configurable in the OS1-64 (varying the vertical FoV). Livox lidars have a non-repetitive scan pattern that delivers higher angular resolution with longer integration times. For lidars, range is based on manufacturer information, with values corresponding to 80\% Lambertian reflectivity and 100 klx sunlight, except for the L515 lidar camera.} 
    \renewcommand{\arraystretch}{1.23}
    \begin{tabular}{@{}lcccccccc@{}}  
        \toprule
        & IMU & Type & Channels & FoV & Angular Resolution & Range & Freq. & Points   \\
        \midrule   
        \textbf{Velodyne VLP-16} & N/A & spinning & 16  & 360°×30°    & V:2.0°, H:0.4°    & 100\,m      & 10\,Hz  & 300,000 pts/s \\   
        
        \textbf{Ouster OS1-64}      &  ICM-20948 & spinning & 64 & 360°×45°   & V:0.7°, H:0.18°    & 120\,m       & 10\,Hz  & 1,310,720 pts/s \\  
 
        \textbf{Ouster OS0-128}      & ICM-20948 & spinning & 128  & 360°×90°    & V:0.7°, H:0.18°    & 50\,m       & 10\,Hz  & 2,621,440 pts/s \\  
 
       \textbf{Livox Horizon}   &  BOSCH BMI088 & solid-state & N/A & 81.7°×25.1°          & N/A                 & 260\,m      & 10\,Hz  & 240,000 pts/s \\  
        
        \textbf{Livox Avia}      &  BOSCH BMI088 & solid-state & N/A & 70.4°×77.2°         & N/A                 & 450\,m      & 10\,Hz  & 240,000 pts/s\\  
 
        \textbf{RealSense L515}  &   BOSCH BMI085 & lidar camera & N/A & 70°×43°($\pm 3$°)          & N/A       & 9\,m        & 30\,Hz  & -  \\
        \textbf{RealSense T265}  & BOSCH BMI055 & fisheye cameras & N/A & 163±5°          & N/A       & N/A      & 30\,Hz  & -  \\
     \bottomrule
    \end{tabular}
     \label{table:sensor_details}
\end{table*} 

\begin{table}  
    \centering
    \caption{List of data sequences in our extended dataset. The table includes the sequences introduced in our previous work~\cite{li2022multi}, together with new sequences showcasing new ground truth data sources. The five lidars indicated (5x Lidars) and cameras are listed in Table~\ref{table:sensor_details}.} 
    \label{table:data_sequences}
    \footnotesize
    \renewcommand{\arraystretch}{1.23}
    \begin{tabular}{@{}llcl@{}}  
        \toprule 
        Sequence        & Description                           & Ground Truth      & \hspace{0.5em}Sensor setup \\
        \midrule   
        Forest01-03  & Previous dataset~\cite{li2022multi}      & MOCAP/ SLAM       & \multirow{3}{2.3em}{$\left\{
                    \begin{tabular}{l}
                        5x Lidars \\
                        L515 \\
                        Optitrack
                    \end{tabular}\right.$} \\
        Indoor01-05     & Previous dataset~\cite{li2022multi}   & MOCAP/ SLAM       &  \\
        Road01-02       & Previous dataset~\cite{li2022multi}   & SLAM              &  \\
        Indoor06        & Lab space (easy)                      & MOCAP             &  \multirow{7}{2.3em}{$\left\{
                    \begin{tabular}{l}
                        \\
                        5x Lidars \\
                        L515 \\
                        T265 \\
                        Optitrack \\
                        GNSS \\
                        \\
                    \end{tabular}\right.$} \\ 
        Indoor07        & Lab space (hard)                      & MOCAP             &    \\
        Indoor08        & Classroom space                       & SLAM+ICP          &    \\
        Indoor09        & Corridor (short)                      & SLAM+ICP          &    \\
        Indoor10        & Corridor (long)                       & SLAM+ICP          &    \\
        Indoor11        & Hall (large)                          & SLAM+ICP          &    \\
        Road03          & Open road                             & GNSS RTK          &    \\
        \bottomrule
    \end{tabular}
    \vspace{1em}
\end{table}


\section{Related Works}\label{sec:related_works}

Owing to high accuracy, versatility, and resilience across environments, 3D lidar SLAM has received much study as a crucial component of robotic and autonomous systems~\cite{cadena2016past}. In this section, we limit the scope to the well-known and well-tested 3D lidar SLAM methods. We also include an overview of the most recent 3D lidar SLAM benchmarks.

\subsection{3D Lidar SLAM} \label{subsec:3d_lidar_slam}
The primary types of 3D lidar SLAM algorithms today are lidar-only~\cite{rozenberszki2020lol}, and loosely-coupled~\cite{zhen2017robust} or tightly-coupled~\cite{ye2019tightly} with IMU data. Tightly-coupled approaches integrate the lidar and IMU data at an early stage, in opposition to SLAM methods that loosely fuse the lidar and IMU outputs towards the end of their respective processing pipelines.

In terms of lidar-only methods, an early work by Zhang et al. on Lidar Odometry and Mapping (LOAM) introduced a method that can achieve low-drift and low-computational complexity already in 2014~\cite{zhang2014loam}. Since then, there have been multiple variations of LOAM that enhance its performance. By incorporating a ground point segmentation and a loop closure module,  LeGO-LOAM is more lightweight with the same accuracy but improved computational expense and lower long-term drift~\cite{shan2018lego}. However, lidar-only approaches are mainly limited by a high susceptibility to featureless landscapes~\cite{li2020localization, nevalainen2022long}. By incorporating IMU data into the state estimation pipeline, SLAM systems naturally become more precise and flexible.

In LIOM~\cite{ye2019tightly}, the authors proposed a novel tightly-coupled approach with lidar-IMU fusion based on graph optimization which outperformed the state-of-the-art lidar-only and loosely coupled. Owing to the better performance of tightly-coupled approaches, subsequent studies have focused in this direction. Another practical tightly-coupled method is Fast-LIO~\cite{xu2021fast}, which provides computational efficiency and robustness by fusing the feature points with IMU data through a iterated extended Kalman filter. By extending FAST-LIO, FAST-LIO2~\cite{xu2022fast} integrated a dynamic structure ikd-tree to the system that allows for the incremental map update at every step, addressing computational scalability issues while inheriting the tightly-coupled fusion framework from FAST-LIO.

The vast majority of these algorithms function well with spinning lidars. Nonetheless, new approaches are in demand since new sensors such as solid-state Livox lidars have emerged novel sensing modalities, smaller FoVs and irregular samplings have emerged~\cite{li2022multi}. Multiple existing studies using enhanced SLAM algorithms are being researched to fit these new lidar characteristics. Loam livox ~\cite{lin2020loam} is a robust and real-time LOAM algorithm for these types of lidars. LiLi-OM~\cite{li2021towards} is another tightly-coupled method that jointly minimizes the cost derived from lidar and IMU measurements for both solid-state Lidars and conventional Lidars.

It is worth mentioning that there are other studies addressing lidar odometry and mapping by fusing not only IMU but also visual information or other ranging data for more robust and accurate state estimation~\cite{lin2022r, nguyen2021viral}. 

\subsection{SLAM benchmarks}\label{subsec:3d_lidar_slam_benchmark}
There are various multi-sensor datasets available online. We had a systematic comparison of the popular datasets in the Table \RomanNumeralCaps{3} of our former work~\cite{li2022multi}. Among these datasets, not all of them have an analytical benchmark of 3D Lidar SLAM based on multi-modality Lidars. KITTI benchmark~\cite{geiger2013kitti} is the most significant one with capabilities of evaluating several tasks including odometry, SLAM, objects detection, tracking ans so alike.

 \section{Data Collection}
    
    Our data collection platform is shown in Fig.~\ref{fig:hardware_cfg}, and details of sensors are listed in Table~\ref{table:sensor_details}. The platform has been mounted on a mobile wheeled vehicle to adapt to varying environments. In most scenarios, the platform is manually pushed or teleoperated, except for the forest environment where the platform is handheld.
 
    
\subsection{Data Collection Platform}
     
    The data collection platform contains various lidar sensors, from traditional  spinning lidars with different resolutions to novel solid-state lidar featured with non-repetition scanning patterns. A lidar camera and stereo fisheye camera are also included.
    There are three spinning lidars, a 16-channel Velodyne lidar (VLP-16), a 64-channel Ouster lidar (OS1), and a 128-channel Ouster lidar (OS0). The OS0 and OS1 sensors were mounted left and right sides, where the OS1 is turned 45 degrees clockwise, and the OS0 is turned 45 degrees anticlockwise. The Velodyne lidar is at the top-most position.
    Two solid-state lidars, Horizon and Avia, were installed in the center of the frame. The Optitrack marker set for the MOCAP-based and the antenna for GNSS/RTK ground truth are both fixed on the top of the aluminum stick to maximize its visibility and detection range.
    All sensors are connected to a computer, featuring an Intel i7-10750h processor, 64\,GB of DDR4 RAM memory and 1\,TB SSD storage, through a Gigabit Ethernet router. The data collection system, including sensor drivers and online calibration scripts, are running on ROS Melodic under Ubuntu 18.04 entirely owing to the wider variety of ROS-based lidar SLAM methods available for Melodic.

\subsection{Calibration and Synchronization}
\begin{figure}  
     \centering
     \includegraphics[width=0.48\textwidth]{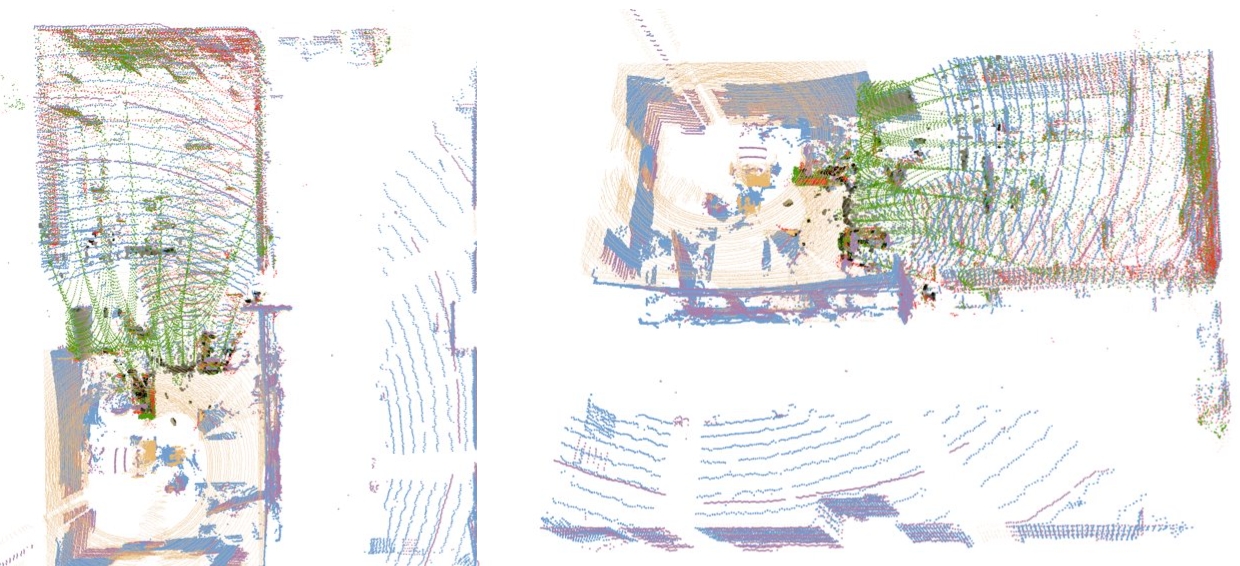}
    \caption{Top view of pointcloud data generated for the calibration process with multiple lidars. The red and green pointclouds represent data obtained from the Livox Horizon and Avia, respectively. The pruple, yellow, blue and black clouds are from the VLP-16, OS1, OS0 and L515 sensors, respectively.} 
    \label{fig:extrin_param}
\end{figure}

     Efficient extrinsic parameters calibration is crucial to multi-sensor platforms, especially for handmade devices  where the extrinsic parameters may change due to unstable connections or distortion of the material during transit. Similar to our previous work~\cite{li2022multi}, we calculated the extrinsic parameter of sensors before each data collecting process. Fig.~\ref{fig:extrin_param} shows the calibration result of sample lidar data from one of the indoor data sequences.
  
Different to our previous work~\cite{li2022multi}, where the timestamp of Ouster and Livox lidars are kept based on their own clock, we synchronized all lidar sensors in ethernet mode via the software-based precise timestamp protocol (PTP)~\cite{lixia2012software}. We compared the orientation estimation between the sensor's built IMUs, and SLAM results with lidars and concluded that the latency of our system is below 5\,ms.

\begin{figure}
    \centering
    \includegraphics[width=.48\textwidth]{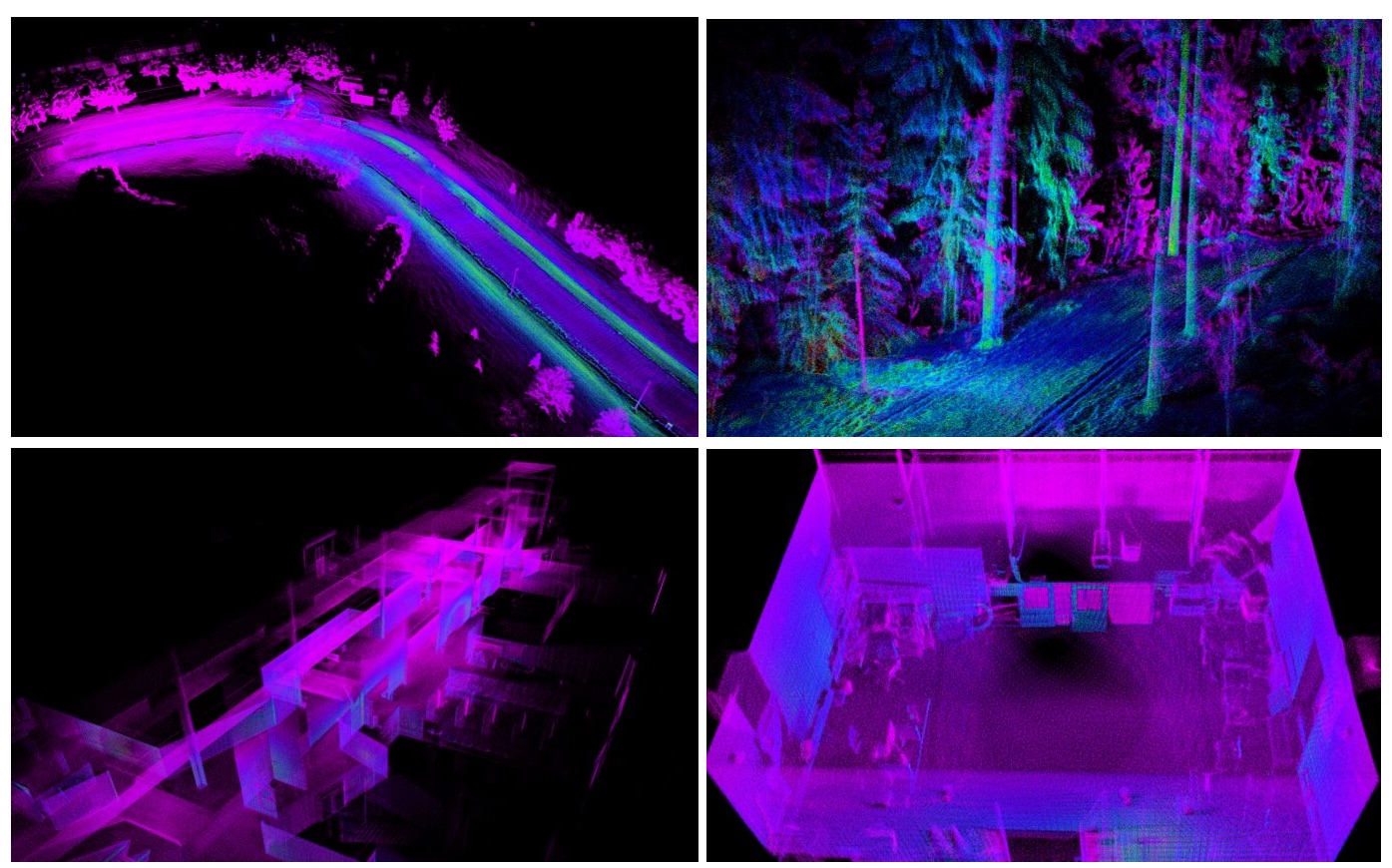}
    \caption{Samples of map data form different dataset sequences. From left to right and top to down, we display maps generated from a forest, an urban area, an open road, and a large indoors lab space, respectively.}
    \label{fig:dataset_env}
\end{figure}

\begin{figure}[t] 
    \centering   
    \includegraphics[width=0.4\textwidth]{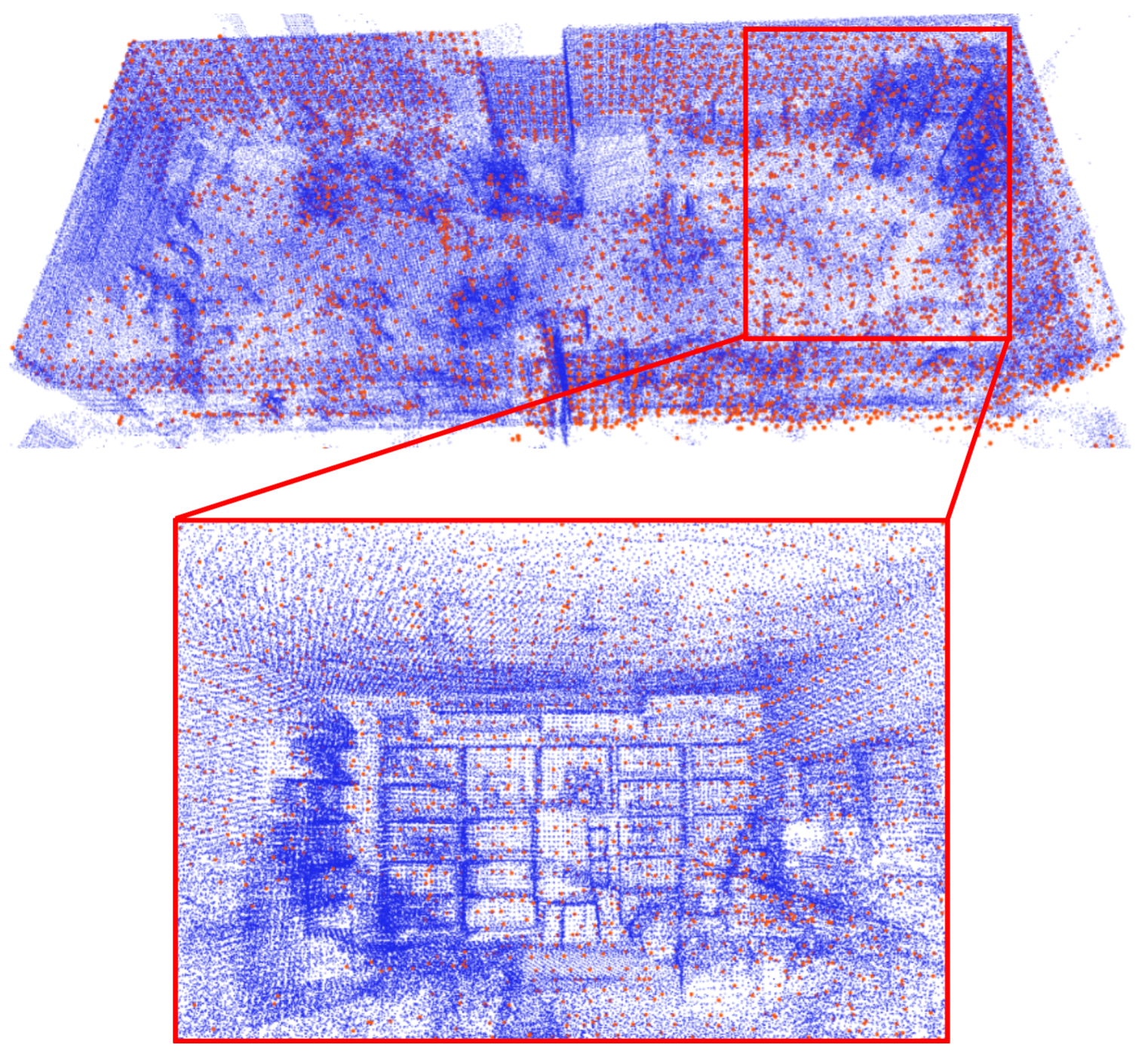}  
    \caption{ NDT localization with ground truth map. External view and Internal view when the current laser scan (orange) is aligned with the Ground truth map (blue).}
    \label{fig:ndt_details} 
\end{figure}

\subsection{SLAM assisted Ground Truth Map}   
\label{icp-gd}
    To provide accurate ground truth for large-scale indoor and outdoor environments, where the MOCAP system is unavailable or GNSS/RTK positioning result becomes unreliable due to the multi-path effect, we propose a SLAM-assisted solid-state lidar-based ground map generation framework.
    
    Inspired by the prior map generation methods in ~\cite{ramezani2020colleage}, where a survey-grade 3D imaging laser scanner Leica BLK360 scanner is unitized to obtain static pointclouds of the target environment, we employed a low-cost solid-state lidar Livox Avia and high resolution spinning lidar to collect undistorted pointclouds from environments. 
    According to the Livox Avia datasheet,
    the range accuracy of the Avia sensor is 2\,cm with a maximum detection range of 480\,m. Due to the non-repetitive scanning pattern, the environment coverage of the pointcloud within the FoV increases with time. Therefore, we integrated multiple frames when the platform is stationary to get more detailed undistorted environmental sampling. Each integrated pointcloud contains more than 240,000 points. The Livox built-in IMU is used to detect the stationary state of the platform when the acceleration values are smaller than 0.01 $m/s^2$ along all axes. After gathering multiple undistorted pointcloud submaps from the target environment, the next step is to match and merge all submap into a global map by ICP.  As the ICP process requires a good initial guess, we employ a high resolution spinning lidar os0 with a 360-degree horizontal FOV to provide raw position by performing real-time SLAM algorithms. This process is outlined in Algorithm~\ref{alg:prior_map}. A dense and high-definition ground truth map can be obtained by denoising the map generated by the algorithm described above to remove noise. Fig.~\ref{fig:ground_ground_map} shows ground truth map of sequence indoor08 generated based on Algorithm~\ref{alg:prior_map}
    
    Let $\mathcal{P}_{sk}$ be the pointcloud produced by the spinning lidar, $\mathcal{P}_{dk}$ be the pointcloud generated by solid-state lidar, and $\mathcal{I}_{k}$ be the IMU data from built-in IMU. 
    Our previous work has shown high resolution spinning lidar has the most robust performance in diverse environments. Therefore, LeGo-LOAM~\cite{shan2018lego} is performed with a high resolution spinning lidar (OS0-128) and outputs the estimated pose for each submap. 
    
    The cached data $\mathcal{S}_{cache}$ stores submaps and the related poses. Let $\mathcal{P}_i$ be the pointcloud and related pose $\textbf{p}_i$ in $\mathcal{S}_{cache}[i]$. 
    The submap $\mathcal{P}_i$ will be first transformed to map coordinate as $\mathcal{P}^m_i$ based on estimated pose $\textbf{p}_i$; 
    then GICP methods are employed on $\mathcal{P}^m_i$ to minimize the Euclidean distance between closest points against pointcloud $\mathcal{M}_{ap}$ iteratively; 
    $\mathcal{P}^m_i$  will be transformed by the transformation matrix generated from GICP process, then merged to the map $\mathcal{M}_{ap}$. The result map $\mathcal{M}_{ap}$ is treated as ground truth map.
   
    After the ground truth map generated, we employ normal NDT method in ~\cite{biber2003normal} to match the real-time pointcloud data from spinning lidar against the HR map as the Figure~\ref{fig:ndt_details} shows to get the platform position in ground truth map. The matching result from the NDT localizer is treated as the ground truth.


\begin{figure}[t]
    \centering
    \setlength{\figurewidth}{0.48\textwidth}
    \setlength{\figureheight}{0.48\textwidth}
    \scriptsize{\input{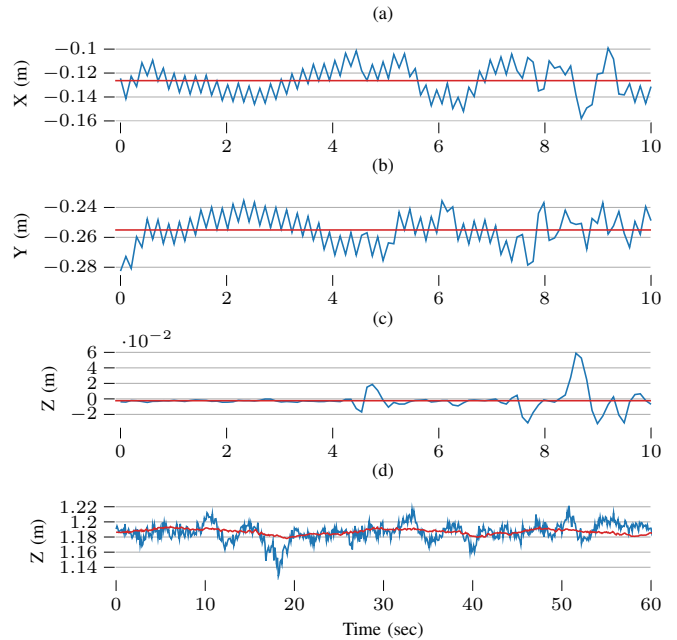}}
    \caption{ \scriptsize{\textbf{(a) (b) (c):} Ground truth position values for the first 10 seconds of the dataset when the device was stationary. Red lines show the mean values over this period of time. \textbf{(d):} Comparison of NDT-based ground-truth z-values (green) to MOCAP-based z-values (red)  over the course of 60 seconds of the dataset while the device was in motion. }}
    \label{fig:gt_still_var}
\end{figure}


 
\begin{algorithm}[t]
    \footnotesize
	\caption{SLAM-assisted ICP-based prior map generation for ground truth data.} 
	\label{alg:prior_map}
	\KwIn{ \\
	    \vspace{.42em}
	    \hspace{1em}Spinning lidar pointcloud:  $\mathcal{P}_{sk} $ \\
	    \vspace{.23em}
	    \hspace{1em}Solid-state lidar pointcloud:  $\mathcal{P}_{sk} $ \\
	    \vspace{.23em}
	    \hspace{1em}IMU data:                    $\mathcal{I}_{k}$ \\
	    \vspace{.23em}
	}
	\KwOut{  \\
	    \vspace{.42em}
	    \hspace{1em}Platform state: $\textbf{p}_{k}$ \\
	    \vspace{.23em}
	    \hspace{1em}Prior map: $\mathcal{M}_{ap} $
	}  
	\BlankLine 
	\While{new $\mathcal{P}_{sk}$}{
	    \vspace{.23em}
	    $\textbf{p}^{k}  \leftarrow SLAM ( \mathcal{P}_{sk}$)\;
	    \vspace{.23em}
	}
	\BlankLine 
	\tcp{Cached still clouds and raw pose}
	$\mathcal{S}_{cache} = \{\}$\;
	\vspace{.42em}
	\tcp{Cached still cloud }
	$\mathcal{P}_{cache} = []$\;  
	\vspace{.42em}
	
    \While{new $\mathcal{P}_{dk}$ }
    { 
        \eIf{ $\mathcal{I}_{k}.V_{angular} < th_a$, $\textbf{p}_{k}.V_{linear} < th_v$}
            { 
                \vspace{.23em}
                $s = True$\;
    	        \vspace{.23em}
                $\mathcal{P}_{m} = \mathcal{P}_{m} + \mathcal{P}_{dk} $\; 
            }
            { 
                \vspace{.23em} 
                $s = False$\;
    	        \vspace{.23em}
    	        $\mathcal{P}_{cache}.clear()$\;
    	        \vspace{.23em} 
                $\mathcal{S}_{cache} \leftarrow (\mathcal{P}_{m}, \textbf{p}_{k})$\;
            }
    }  
    \BlankLine 
    \While{$\mathcal{S}_{cache}.size()> 0$}{
        \vspace{.23em}$\mathcal{M}_{ap}   \leftarrow  ICP(\mathcal{S}_{cache}, \textbf{p}_{k} ,  \mathcal{M}_{ap} )$\;
        \vspace{.23em}$\mathcal{S}_{cache}.clear()$\;
        \vspace{.23em}
    }
\end{algorithm}

 \begin{table*}[t]
\centering
\caption{ Absolute position error (APE) ($\mu/\sigma$) in \textit{cm} of the selected methods (N/A when odometry estimations diverge). Best results 
in bold. } 
\renewcommand{\arraystretch}{1.6}
\resizebox{\linewidth}{!}{%
\begin{tabular}{@{}lcccccccccc@{}}
\hline
Sequence & FLIO\_OS0 & FLIO\_OS1 & FLIO\_Velo & FLIO\_Avia & FLIO\_Hori & LLOM\_Hori & LLOMR\_OS1 & LIOL\_Hori & LVXM\_Hori & LEGO\_Velo \\ \hline
Indoor06 & \textbf{0.015 / 0.006} & 0.032 / 0.011 & N/A & 0.205 / 0.093 & 0.895 / 0.447 & N/A & 0.882 / 0.326 & N/A & N/A & 0.312 / 0.048 \\
Indoor07 & \textbf{0.022 / 0.007} & 0.025 / 0.013 & 0.072 / 0.031 & N/A & N/A & N/A & N/A & N/A & N/A & 0.301/0.081 \\
Indoor08 & 0.048 / 0.030 & \textbf{0.042 / 0.018} & 0.093 / 0.043 & N/A & N/A & N/A & N/A & N/A & N/A & 0.361 / 0.100 \\
Indoor09 & \textbf{0.188 / 0.099} & N/A & 0.472 / 0.220 & N/A & N/A & N/A & N/A & N/A & N/A & N/A \\
Indoor10 & 0.197 / 0.072 & \textbf{0.189 / 0.074} & 0.698 / 0.474 & 0.968 / 0.685 & 0.322 / 0.172 & 1.122 / 0.404 & 1.713 / 0.300 & 0.641 / 0.469 & N/A & 0.930 / 0.901 \\
Indoor11 & 0.584 / 0.080 & \textbf{0.105 / 0.041} & 0.911 / 0.565 & 0.196 / 0.098 & 0.854 / 0.916 & 0.1.097 / 0.0.45 & 1.509 / 0.379 & N/A & N/A & N/A \\
Road03 & 0.123 / 0.032 & \textbf{0.095 / 0.037} & 1.001 / 0.512 & 0.211 / 0.033 & 0.351 / 0.043 & 0.603 / 0.195 & N/A & 0.103 / 0.058 & 0.706 / 0.396 & 0.2464 / 0.063 \\
Forest01 & 0.138 / 0.054 & 0.146 / 0.087 & N/A & 0.142 / 0.074 & 0.125 / 0.062 & 0.116 / 0.053 & 0.218 / 0.110 & \textbf{0.054 / 0.033} & 0.083 / 0.041 & 0.064 / 0.032 \\
Forest02 & 0.127 / 0.065 & \textbf{0.121 / 0.069} & N/A & 0.211 / 0.077 & 0.348 / 0.077 & 0.612 / 0.198 & N/A & 0.125 / 0.073 & 0.727 / 0.414 & 0.275 / 0.077  \\ \hline
\end{tabular}
}
\label{tab:ate_error}
\end{table*}

 
 
 \begin{table*}[t] 
\centering
\caption{Average run-time resource (CPU/RAM) utilization and performance (pose calculation speed) comparison of selected SLAM methods across multiple platforms. For the pose publishing frequency, the data is played at 15 times the real speed. CPU utilization of 100\% equals one full processor core.}
\renewcommand{\arraystretch}{1.8}
\resizebox{1\linewidth}{!}{%
\begin{tabular}{@{}lcccccccccc@{}}
    
\hline
    & \multicolumn{10}{c}{( CPU utilization (\%), RAM utilization (MB), Pose publication rate (Hz) )} \\[-.42em]
 & FLIO\_OS0 & FLIO\_OS1 & FLIO\_Velo & FLIO\_Avia & FLIO\_Hori & LLOM\_Hori                        & LLOMR\_OS1 & LIOL\_Hori & LVXM\_Hori & LEGO\_Velo \\ 
  \hline
\textbf{Intel PC}   & (79.4, 384.5, 74.0)    & (73.7, 437.4, 67.5)   & (69.9, 385.2, 98.6)      & (65.0, 423.8, 98.3)     & (65.7, 423.8, 103.7)     & \multicolumn{1}{l}{(126.2, 461.6, 14.5)} & (112.3, 281.5, 25.8)    & (186.1, 508.7, 19.1)    & (135.4, 713.7, 14.7)     & (28.7, 455.4, 9.8)     \\
\textbf{AGX MAX}     & (40.9, 385.3, 13.6)  & (54.5, 397.5, 21.2)  & (44.4, 369.7, 29.1)   & (40.8, 391.5, 32.3) & (37.6, 408.4, 34.7)  & \multicolumn{1}{l}{(128.5, 545.4, 9.1)} & (70.8, 282.3, 9.6)  & (247.2, 590.3, 9.6)    & (162.3, 619.0, 10.5)   & (42.4, 227.8, 7.0)  \\ 
\textbf{AGX 30\,W} & (55.1, 398.8, 13.2)  & (73.9, 409.2, 15.4)  & (58.3, 367.6, 21.4)   & (47.4, 413.4, 24.5) & (50.5, 387.9, 26.8)  & \multicolumn{1}{l}{(168.5, 658.5, 1.5)} & (107.1, 272.2, 6.5)  & (188.1, 846.0, 4.1)    & (185.86, 555.81, 5.0)   & (62.8, 233.4, 3.5)  \\ 
\textbf{UP Xtreme}   & (90.9, 401.8, 47.3)    & (125.9, 416.2, 58.0)   & (110.5, 380.5, 89.6)      & (113.2, 401.2, 90.7)     & (109.7, 422.8, 91.0)     & \multicolumn{1}{l}{(130.1, 461.1, 12.8)} & (109.0, 253.5, 13.6)    & (298.2, 571.8, 14.0)    & (189.6, 610.4, 7.9)     & (39.7, 256.6, 9.1)     \\
\textbf{NX 15\,W}    & (53.7, 371.1, 14.3)    & (73.3, 360.4, 14.2)   & (57, 331.5, 19.5)      & (51.2, 344.8, 21.9)     & (47.5, 370.7, 23.4)     & \multicolumn{1}{c}{ (N / A)} & (N / A)    & (239.0, 750.5, 4.54)    & (198.0, 456.7, 5.5)     & (36.9, 331.4, 3.7)     \\ \hline
\end{tabular}%
}
\label{tab:runtime_src}
\end{table*}
 
  
\section{SLAM benchmark}\label{sec:methodology}
In this study, we evaluated popular 3D Lidar SLAM algorithms in multiple data sequences of various scenarios, including indoor, outdoor, and forest environments. 

\subsection{Ground Truth Evaluation} 

The evaluation of the accuracy of the proposed ground truth prior map method is challenging for some scenes in the dataset, as both GNSS and MOCAP systems are not available in indoor environments such as long corridors.
Figure~\ref{fig:gt_still_var} (a),(b),(c) shows the standard deviations of the ground truth generated by the proposed method during the first 10 seconds when the device is stationary from sequence \textit{Indoors09}. The standard deviations along the $X$, $Y$, and $Z$ axes are 2.2\,cm, 4.1\,cm, and 2.5\,cm, respectively, or about 4.8\,cm overall. 
However, evaluating localization performance when the device is in motion is more difficult. To better understand the order of magnitude of the accuracy, we compare the NDT-based ground truth $Z$ values with the MOCAP-based ground truth $Z$ values in the sequence \textit{Indoor06} environment. The results in Fig. ~\ref{fig:gt_still_var} (d) show that the maximum difference does not exceed 5\,cm. 

\subsection{Lidar Odometry Benchmarking}
     Different types of SLAM algorithms are selected and tested in our experiment.
     Lidar-only algorithms LeGo-LOAM (LEGO)
    ~\footnote{\href{https://github.com/RobustFieldAutonomyLab/LeGO-LOAM}{https://github.com/RobustFieldAutonomyLab/LeGO-LOAM}} and Livox-Mapping (LVXM)~\footnote{\href{https://github.com/Livox-SDK/livox_mapping}{https://github.com/Livox-SDK/livox\_mapping}} are applied on data from the VLP-16 and Horizon separately; 
    Tightly-coupled iterated extended Kalman filter-based methods, FAST-LIO (FLIO)~\footnote{\href{https://github.com/hku-mars/FAST_LIO}{https://github.com/hku-mars/FAST\_LIO}} ~\cite{xu2021fastlio}, are applied on both spinning lidar and solid-state lidar with built-in IMUs; 
    A tightly coupled lidar inertial SLAM system based on sliding window optimization, LiLi-OM~\footnote{\href{https://github.com/KIT-ISAS/lili-om}{https://github.com/KIT-ISAS/lili-om}}~\cite{li2020towards} is tested with  OS1 and Horizon.
    Furthermore,  a tightly coupled method featuring sliding window optimization developed for Horizon lidar, LIO-LIVOX (LIOL)\footnote{\href{https://github.com/Livox-SDK/LIO-Livox}{https://github.com/Livox-SDK/LIO-Livox}} has also been tested on Horizon lidar data. 

    We provide a quantitative analysis of the odometry error based on the ground truth in Table~\ref{tab:ate_error}. To compare the trajectories in the same coordinate, we treat the coordinate of OS0 as a reference coordinate and transformed all trajectories generated by selected SLAM methods to reference coordinate. The absolute pose errors (APE)~\cite{sturm2012benchmark} is employed as the core evaluation metric. We calculated the error of each trajectory with the open-source EVO toolset\,\footnote{\href{https://github.com/MichaelGrupp/evo.git}{https://github.com/MichaelGrupp/evo.git}}. 
    
    From the result, we can conclude that FAST\_LIO with high resolution spinning lidar OS0 and OS1 has the most robust performance that can complete all the trajectories on different sequences with promising accuracy.  
    Especially for sequence \textit{Indoor09}  showcasing a long corridor, all other methods failed and Fast\_LIO with high resolution lidar remain survived. Solid-state lidar-based SLAM systems such as LIOL\_Hori perform as well or even better in outdoor environments than rotating lidars with appropriate algorithms, but perform significantly more poorly in the indoor environments.
    For the open road sequences \textit{Road03}, all SLAM methods perform well, and the trajectories are completed without major disruptions.  
%
    For the indoor sequence \textit{Indoor06}, Avia-based and Horizon-based FLIO are able to reconstruct the sensor trajectory but significant drift accumulates.
    In all of these sequences, all the methods applied to spinning lidars perform satisfactorily. This result can be expected as they have full view of the environment, which has a clear geometry. 
    For the sequence \textit{Indoor10} showcasing a long corridor, almost all methods can reconstruct the complete trajectory again. The best performance comes from OS0-FLIO and OS1-FLIO with correct alignment between the first and last positions. We hypothesize that this occurs because OS0 has more channels than OS1, leading to lower accumulated cumulative angular drift.
 
\begin{figure*}[t]
    \centering
        \begin{subfigure}{.32\textwidth}
            \centering
            \includegraphics[width=\textwidth]{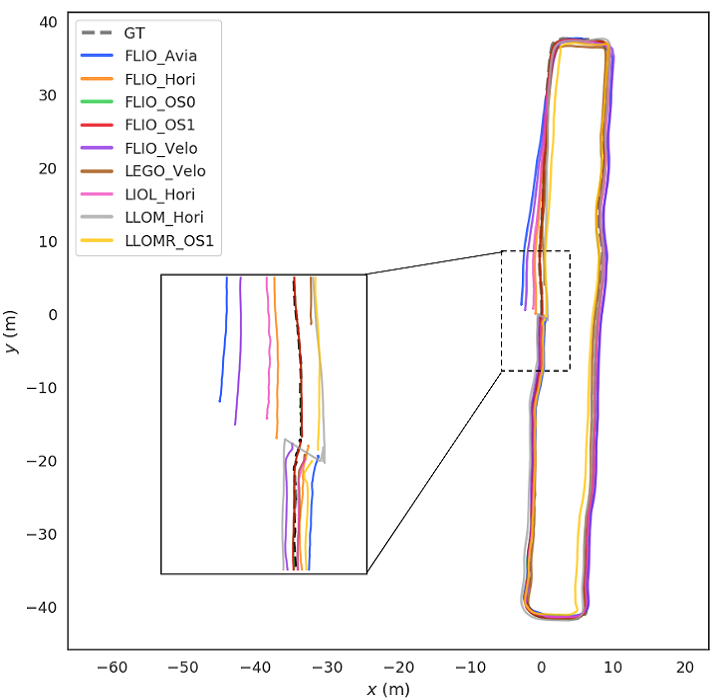}
            \caption{trajectory comparison of sequence \textit{Indoor10} } 
            \label{fig:3rd_traj}
        \end{subfigure}
    \hfill
        \begin{subfigure}{.32\textwidth}
          \centering
            \includegraphics[width=\textwidth]{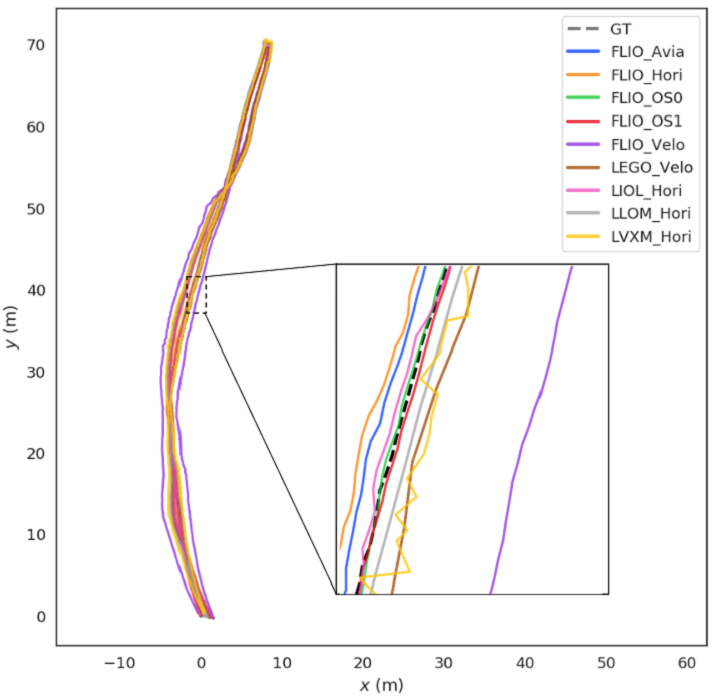}
            \caption{trajectory comparison of sequence \textit{Road03} }
          \label{fig:road_traj}
        \end{subfigure}
    \hfill
        \begin{subfigure}{.32\textwidth}
          \centering
            \includegraphics[width=\textwidth]{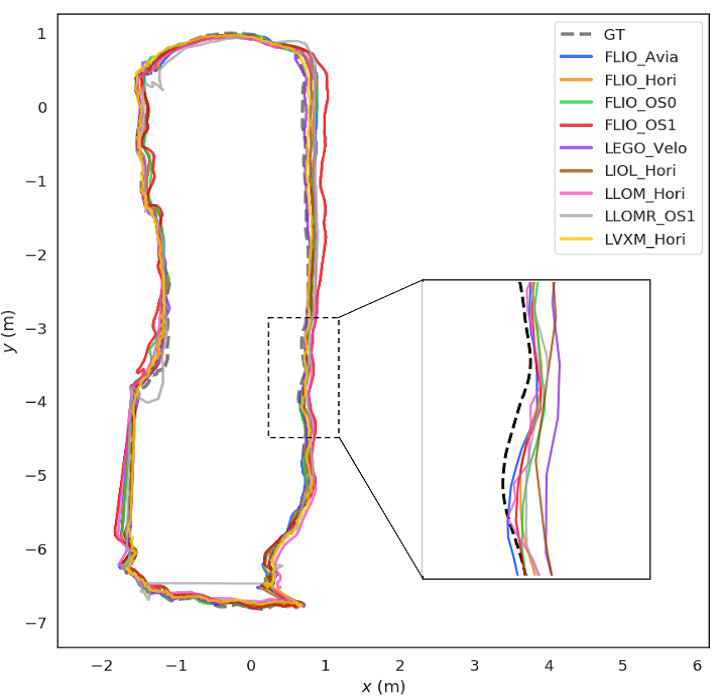}
            \caption{trajectory comparison of sequence \textit{Forest01} } 
          \label{fig:resnet_real_acc_mixed}
        \end{subfigure}
    \caption{Demos of trajectories generated by multiple 3D LiDAR SLAM based on data from indoor, road, and wild environments}
    \label{fig:demo_traj}
    \vspace{-1em}
\end{figure*}

    In addition to the quantitative trajectory analysis, we visualize trajectories generated by selected methods in 3 representative environments (indoors, outdoors, forest) in Fig.~\ref{fig:demo_traj}. 
    Full reconstructed paths are available in the dataset repository. 
 
 
\subsection{Run-time evaluation across certain computing platforms}
We conducted this experiment on 4 different platforms. First, a Lenovo Legion Y7000P with 16\,GB RAM, a 6-core Intel i5-9300H (2.40\,GHz) and an Nvidia GTX 1660Ti (1536 CUDA cores, 6\,GB VRAM). Then, the Jetson Xavier AGX, a popular computing platform for mobile robots, has an 8-core ARMv8.2 64-bit CPU (2.25\,GHz), 16\,GB RAM and 512-core Volta GPU. From its 7 power modes, we chose MAX and 30\,W (6 core only) modes. The Nvidia Xavier NX is also a common embedded computing platform with a 6-core ARM v8.2 64-bit CPU, 8\,GB RAM, and 384-core Volta GPU with 48 Tensor cores. For the NX, we choose the 15\,W power mode (all 6 cores). Finally, the UP Xtreme board features an 8-core Intel i7-8665UE (1.70\,GHz) and 16\,GB RAM.

These platforms all run ROS Melodic on Ubuntu 18.04. The CPU and memory utilization is measured with a ROS resource monitor tool~\footnote{\href{https://github.com/alspitz/cpu_monitor}{https://github.com/alspitz/cpu\_monitor}}. Additionally, for minimizing the difference of the operating environment,  we unified the dependencies used in each SLAM system into same version, and each hyperparameter in the SLAM system is configured with the default values. The results are shown in Table~\ref{tab:runtime_src}.

The memory utilization of each selected SLAM approach among the two processor architectures platforms are roughly equivalent. However, the CPU utilization of the same SLAM algorithm running on Intel processors is generally higher than the other algorithms, and also the highest publishing frequency is obtained. LeGO\_LOAM has the lowest CPU utilization but its accuracy is towarsd the low end (see Table~\ref{tab:ate_error}), and has a very low pose publishing frequency. Fast-LIO performs well, especially on embedded computing platforms, with good accuracy, low resource utilization, and high pose publishing frequency. In contrast, LIO\_LIVOX has the highest CPU utilization due to the computational complexity of the frame-to-model registration method applied to estimate the pose.

A final takeaway is in the generalization of the studied methods. Many state-of-the-art methods are only applicable to a single lidar modality. In addition, those that have higher flexibility (e.g., FLIO) still lack the ability to support a point-cloud resulting from the fusion of both types of lidars.



\section{Conclusion}\label{sec:conclusions}

In this paper, we provide lidar datasets covering the characteristics of various environments (indoor, outdoor, forest), and systematically evaluate 5 open source SLAM algorithms in terms of lidar Odometry, and power consumption. The experiments have covered 9 sequences across 2 computing platforms.
By including the Nvidia Jetson Xavier platform, it provides further references for the application of various SLAM algorithms on computationally resource-constrained devices such as drones.
Overall, we found that in both indoor and outdoor environments, the spinning lidar-based FLIO exhibited good performance with low power consumption, which we believe is due to the ability of the spinning lidar to obtain a full view of the environment . However, in the forest environment, the LIOL algorithm based on solid-state lidar has the best performance in terms of accuracy and mapping quality, although it has the highest power consumption due to the sliding window optimization. 

Finally, we aim to further extend our dataset to provide more refined and difficult sequences and open source it in the future. In this paper, our benchmark tests only focus on SLAM algorithms based on spinning lidar and solid-state lidar. In the future, we will add benchmark tests based on cameras and even SLAM algorithms based on multiple sensor fusions.



\section*{Acknowledgment}

This research work is supported by the Academy of Finland's AeroPolis project (Grant 348480).

\newpage
\bibliographystyle{IEEEtran}
\bibliography{bibliography}

\end{document}